\documentclass[10pt,twocolumn,letterpaper]{article}
\usepackage{appendix}
\usepackage{cvpr}
\usepackage{times}
\usepackage{epsfig}
\usepackage{graphicx}
\usepackage{amsmath}
\usepackage{amssymb}
\usepackage{amsfonts}
\usepackage{float} 
\usepackage{subfigure}
\usepackage{color}
\usepackage{multirow}
\usepackage{rotating}

\usepackage[T1]{fontenc}
\usepackage[utf8]{inputenc}
\usepackage{authblk}
\usepackage{comment}

\usepackage[breaklinks=true,bookmarks=false]{hyperref}

\cvprfinalcopy 

\def\cvprPaperID{****} 

\begin{document}

\title{DG-Font: Deformable Generative Networks for Unsupervised Font Generation}



\author{{Yangchen Xie \quad Xinyuan Chen\thanks{Corresponding author: xychen@cee.ecnu.edu.cn} } \quad Li Sun \quad Yue Lu\\
{Shanghai Key Laboratory of Multidimensional Information Processing, } \\ 
{East China Normal University, 200241 Shanghai, China} }

\maketitle

\thispagestyle{empty}

\begin{abstract}
Font generation is a challenging problem especially for some writing systems that consist of a large number of characters and has attracted a lot of attention in recent years. However, existing methods for font generation are often in supervised learning. They require a large number of paired data, which is labor-intensive and expensive to collect. Besides, common image-to-image translation models often define style as the set of textures and colors, which cannot be directly applied to font generation. To address these problems, we propose novel deformable generative networks for unsupervised font generation (DG-Font). We introduce a feature deformation skip connection (FDSC) which predicts pairs of displacement maps and employs the predicted maps to apply deformable convolution to the low-level feature maps from the content encoder. The outputs of FDSC are fed into a mixer to generate the final results. Taking advantage of FDSC, the mixer outputs a high-quality character with a complete structure. To further improve the quality of generated images, we use three deformable convolution layers in the content encoder to learn style-invariant feature representations. Experiments demonstrate that our model generates characters in higher quality than state-of-art methods. The source code is available at https://github.com/ecnuycxie/DG-Font.

\end{abstract}

\section{Introduction}

\begin{figure}[ht]
\begin{center}
   \includegraphics[width=0.9\linewidth]{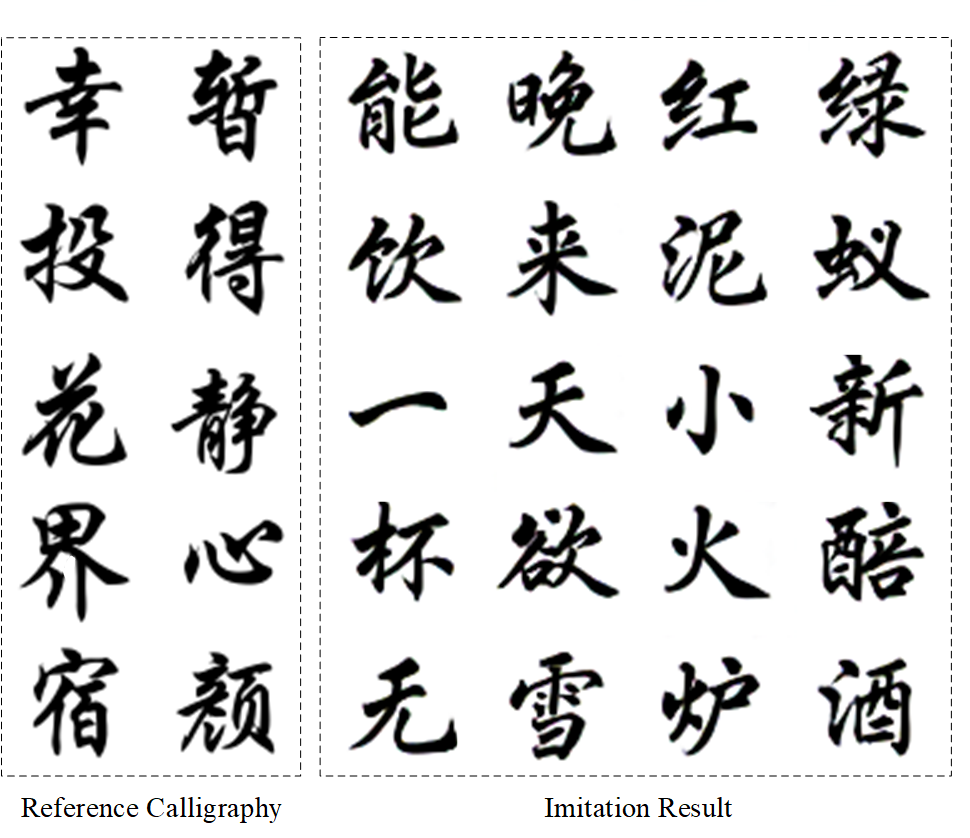}
\end{center}
   \caption{\textbf{Unsupervised font generation results.} The reference calligraphy is a Tang poem written by a calligrapher, and imitation result is another famous Tang poem generated from our model which are with rich details, such as stroke tips, joined-up writing, and thickness of strokes.}   
\label{fig:1}
\end{figure}

Every day, people consume a massive amount of texts for information transfer and storage. As the representation of texts, the font is closely related to our daily life. Font generation is critical in many applications, \eg, font library creation, personalized handwriting, historical handwriting imitation, and data augmentation for optical character recognition and handwriting identification. Traditional font library creating methods heavily rely on expert designers by drawing each glyph individually, which is especially expensive and labor-intensive for logographic languages such as Chinese (more than 60,000 characters), Japanese (more than 50,000 characters), and Korean (11,172 characters). 

Recently, the development of convolutional neural networks enables automatic font generation without human experts. There have been some attempts to explore font generation and achieve promising results. \cite{DBLP:journals/corr/UpchurchSB16,DBLP:conf/cvpr/AzadiFKWSD18,DBLP:conf/cvpr/FogelACML20} utilize deep neural networks to generate entire sets of letters for certain alphabet languages. Two notable projects, ``Rewrite" \cite{Rewrite} and ``zi2zi" \cite{zi-2-zi}, generate logographic language characters by learning a mapping from one style to another with thousands of paired characters. After that, EMD \cite{EMD} and SA-VAE \cite{Sun2018LearningTW} design neural networks to separate the content and style representation, which can extend to generate character of new styles or contents. However, these methods are in supervised learning and required a large amount of paired training samples.

Some other methods exploit auxiliary annotations (\eg, strokes, radicals) to facilitate high-quality font generation. For example, \cite{SCFont} utilizes labels for each stroke to generate glyphs by writing trajectories synthesis. \cite{RD-GAN} employ the radical decomposition (\eg, radicals or sub-glyphs) of characters to achieve font generation for certain logographic language. DM-Font \cite{Few-shot} and its improved version LF-Font \cite{DBLP:journals/corr/abs-2009-11042} propose disentanglement strategies to disentangle complex glyph structures, which help capture local details in rich text design. However, these methods rely on prior knowledge and can only apply to specific writing systems. Some labels such as the stroke skeleton can be estimated by algorithms, but the estimation error would decrease the generated quality. Also, these methods still require thousands of paired data and annotated labels for training. Recently, there are some attempts \cite{DBLP:conf/aaai/0006W20,DBLP:conf/wacv/ChangZPM18} for unsupervised font generation. \cite{DBLP:conf/wacv/ChangZPM18} introduces a novel module that transfers the features across sequential DenseNet blocks \cite{DBLP:conf/cvpr/HuangLMW17}. \cite{DBLP:conf/aaai/0006W20} proposes a fast skeleton extraction method to obtain the skeleton of characters, and then utilize the extracted skeleton to facilitate font generation.

For the problem of image-to-image translation, a series of works in unsupervised learning have been proposed by combining adversarial training \cite{Kim_2017_ICML,Yi_2017_ICCV} with consistent constraints \cite{Zhu2017UnpairedIT,DBLP:journals/corr/TaigmanPW16,benaim2017one}. FUNIT \cite{DBLP:conf/iccv/0001HMKALK19} maps an image of a source class to an analogous image of a target class by leveraging a few target class images. They extract the style feature of the target class images and employ adaptive instance normalization (AdaIN) \cite{DBLP:conf/iccv/HuangB17} to combine the content and the style features. However, these image-to-image translation methods cannot be directly applied to font generation tasks. Although consistent constraints preserve the structure of a content image, they still encounter some problems for font generation (\eg, blurry, missing some strokes). Also, they usually define the style as the set of textures and colors. The AdaIN-based methods transfer style by aligning feature statics, which tends to transform texture and color, which is not suitable to transform local style patterns (\eg, geometric deformation) for the font. Moreover, \cite{DBLP:conf/wacv/ChangZPM18,DBLP:conf/aaai/0006W20} achieve unsupervised font generation by learning a mapping between two fonts directly, they also ignore the geometric deformation for the font. To learn the mapping across geometry variations, \cite{ganimorph} introduces a discriminator with dilated convolutions as well as a multi-scale perceptual loss that is able to represent error in the underlying shape of objects. \cite{transgaga} disentangles image space into a Cartesian product of the appearance and the geometry latent spaces.

Compelled by the above observations, we propose a novel deformable generative model for unsupervised font generation (DG-Font). The proposed method is designed to deform and transform the character of one font to another by leveraging the provided images of the target font. The proposed DG-Font separates style and content respectively and then mix two domain representations to generate target characters.  We introduce a feature deformation skip connection (FDSC) which predicts pairs of displacement maps and employs the predicted maps to apply deformable convolution to the low-level feature maps from the content encoder. The outputs of FDSC are fed into a mixer to generate the final results. To distinguish different styles, we train our model with a multi-task discriminator, which ensures that each style can be discriminated independently. In addition, another two reconstruction losses are adopted to constrain the domain-invariant characteristics between generated images and content images. 

The feature deformation skip connection (FDSC) module is used to transform the low-level feature of content images, which preserves the pattern of character (\eg, strokes and radicals). Different from the image-to-image translation problem that defines style as a set of texture and color, the style of font is basically defined as geometric transformation, stroke thickness, tips, and joined-up writing pattern. For two fonts with the same content, they usually have correspondence for each stroke. Taking advantage of the spatial relationship of fonts, the feature deformation skip connection (FDSC) is used to conduct spatial deformation, which effectively ensures the generated image to have complete structures. 

Extensive experiments demonstrate that our model achieves comparable results to the state-of-the-art font generation methods. Besides, results show that our model is able to extend to generate unseen style character.

\begin{figure*}
\begin{center}
\includegraphics[width=0.8\textwidth]{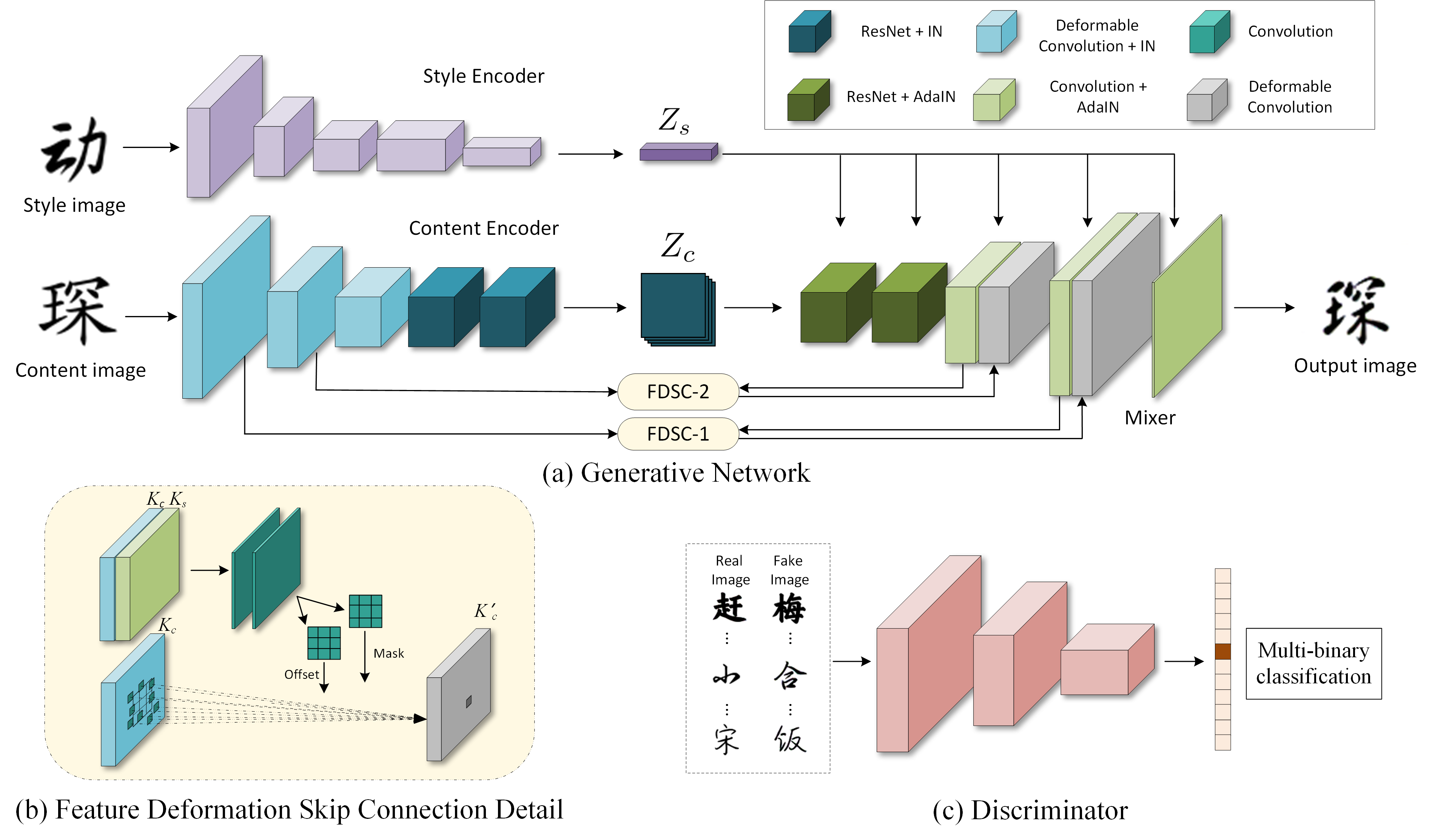}
\end{center}
   \caption{\textbf{Overview of the proposed method}. a) Overview of our generative network. The Style/content encoder maps style/content image to style/content representation $Z_s$/$Z_c$. FDSC-1 and FDSC-2 have the same architecture and apply transformation convolution to the low-level feature from the content encoder and inject the results into the mixer. The mixer generates the output image. b) A detailed illustration of the FDSC module. c) The discriminator output a binary vector, where each element indicates a binary classification to distinguish between generated and real images.}
\label{fig:2}
\end{figure*}

\section{Related works}
\subsection{Image-to-Image Translation}
The purpose of image-to-image translation is to learn a mapping from an image in the source domain to the target domain. Image-to-image translation has been applied in many fields such as artistic style transfer \cite{Johnson2016PerceptualLF,Zhang2018MultistyleGN}, semantic segmentation \cite{Shukla2019ExtremelyWS,Musto2020SemanticallyAI}, image animation \cite{9229197,yang2017pairwise,Chen_2020_CVPR}, object transfiguration \cite{chen2018attention}, and video frames generation \cite{chan2019everybody,9107481,dong2021dual} \etal.
Pix2pix \cite{Isola2017ImagetoImageTW} is the first model proposed for image-to-image translation based on conditional GAN \cite{Mirza2014ConditionalGA}. 
To achieve unsupervised image-to-image translation, a lot of works \cite{Liu2016CoupledGA,Zhu2017UnpairedIT,Bousmalis2017UnsupervisedPD,Shrivastava2017LearningFS} have been proposed, where Cycle-GAN \cite{Zhu2017UnpairedIT} introduces a cycle consistency between source and target domain to discover the relationship of samples between two domain. However, above-mentioned methods can only translate from one domain to another specific domain. To tackle this problem, recent works \cite{DBLP:journals/tip/ChenXYST19,DBLP:conf/iccv/0001HMKALK19,DBLP:journals/corr/abs-2006-06500,DBLP:conf/cvpr/BhattacharjeeKV20} are proposed to simultaneously generate multiple style outputs given the same input. Gated-GAN \cite{DBLP:journals/tip/ChenXYST19} proposes a gated transformer to transfer multiple styles in a single model. FUNIT \cite{DBLP:conf/iccv/0001HMKALK19} encodes content image and class image respectively, and combines them with AdaIN \cite{DBLP:conf/iccv/HuangB17}. TUNIT \cite{DBLP:journals/corr/abs-2006-06500} further introduce a guiding network as an unsupervised domain classifier to automatically produce a domain label of a given image. DUNIT \cite{DBLP:conf/cvpr/BhattacharjeeKV20} extract separate representations for the global image and for the instances to preserve the detailed content of object instances. 
\subsection{Font Generation}
Font generation aims to automatically generate characters in a specific font and create a font library. Recent studies have employed image translation methods for font generation. ``Zi2zi" \cite{zi-2-zi} and ``Rewrite" \cite{Rewrite} implement font generation on the basis of GAN \cite{DBLP:conf/nips/GoodfellowPMXWOCB14} with thousands of character pairs for strong supervision. After that, a series of models are proposed to improve the generated quality based on zi2zi \cite{zi-2-zi}. PEGAN \cite{Sun2018PyramidEG} sets up a multi-scale image pyramid to pass information through refinement connections. HAN \cite{Hierarchical} improves zi2zi by designing a hierarchical loss and skip connection. AEGG \cite{Lyu2017AutoEncoderGG} adds an additional network to refine the training process. DC-Font \cite{Jiang2017DCFontAE} introduces a style classifier to get a better style representation.
However, all the above methods are in supervised learning and require a large number of paired data.

In addition to the paired data, lots of methods employ auxiliary annotations (\eg, stroke and radical decomposition) to further improve the generation quality. SA-VAE \cite{Sun2018LearningTW} disentangles the style and content as two irrelevant domains with encoding Chinese characters into high-frequency character structure configurations and radicals. CalliGAN \cite{CalliGAN} further decomposes characters into components and offers low-level structure information including the order of strokes to guide the generation process. RD-GAN \cite{RD-GAN} proposes a radical extraction module to extract rough radicals which can improve the performance of discriminator and achieves the few-shot Chinese font generation. Also, some other attempts have been made in Chinese character generation by adopting skeleton/stroke extraction algorithm \cite{DBLP:conf/aaai/0006W20,SCFont}. However, they need extra annotations or algorithms to guide font generation; while the estimation error would decrease the generation performance. In contrast, our proposed model, DG-Font, can generate images in an unsupervised way without other annotations.
\subsection{Deformable Convolution}
CNNs have inherent limitations in modeling geometric transformations due to the fixed kernel configuration. To enhance the transformation modeling capability of CNNs, \cite{Deformablev1} proposes the deformable convolutional layer. It augments the spatial sampling locations in the modules with additional offsets. The deformable convolution has been applied to address several high-level vision tasks, such as object detection \cite{Bertasius2018ObjectDI,Deformablev1,Deformablev2} video object detection 
\cite{DBLP:conf/icassp/ChenLF0Y20} sampling, semantic segmentation \cite{Deformablev2}, and human pose estimation \cite{DBLP:conf/eccv/SunXWLW18}. Recently, some methods attempt to apply deformable convolution in the image generation tasks. TDAN \cite{DBLP:conf/cvpr/TianZ0X20} addresses video super-resolution task by using deformable convolution to align two continuous frames and output a high-resolution frame. \cite{DBLP:conf/eccv/YinSL20} synthesizes novel  view images by deformable convolution given the view condition vectors. In our proposed DG-Font, offsets are estimated by a learned latent style code.

\section{Methods}
\subsection{overview}

Given a content image $I_c$ and a style image $I_s$, our model aims to generate the character of the content image with the font of the style image. As illustrated in Fig. \ref{fig:2}, the proposed generative network consists of a style encoder, a content encoder, a mixer, and two feature deformation skip connection (FDSC) modules. The architecture of the style encoder and discriminator is simplified in Fig. \ref{fig:2}. The detailed architecture is shown in Appendix A. \textbf{The style encoder} is designed to learn the style representation from input images. Specifically, the style encoder takes a style image as the input and maps it to a style latent vector $Z_{s}$.
\textbf{The content encoder} is introduced to extract the structure feature of the content images. The content encoder maps the content image into a spatial feature map $Z_{c}$. The content encoder module is made of three deformable convolution layers followed by two residual blocks. The introduced deformable convolution layer enables the content encoder to produce style-invariant features for images with the same content. 
\textbf{The mixer} aims to output characters by mixing the content feature representations $Z_c$ and style feature representations $Z_s$. AdaIN \cite{DBLP:conf/iccv/HuangB17} is adopted to inject the style feature to the mixer. Besides, \textbf{the feature deformation skip connection modules} transfers the deformed low-level feature from the content encoder to the mixer. Details are described in Sec \ref{sec:deform_module}.

When character images are generated from the generative network, \textbf{a multi-task discriminator} is adopted to conduct discrimination for each style simultaneously. For each style, the output of the discriminator is a binary classification whether the input image is a real image or a generated image. As there are several different styles of fonts in the training set, the discriminator outputs a binary vector whose length is the number of styles.

\subsection{Feature Deformation Skip Connection}
\label{sec:deform_module}

\begin{figure}[t]
\begin{center}
   \includegraphics[width=0.9\linewidth]{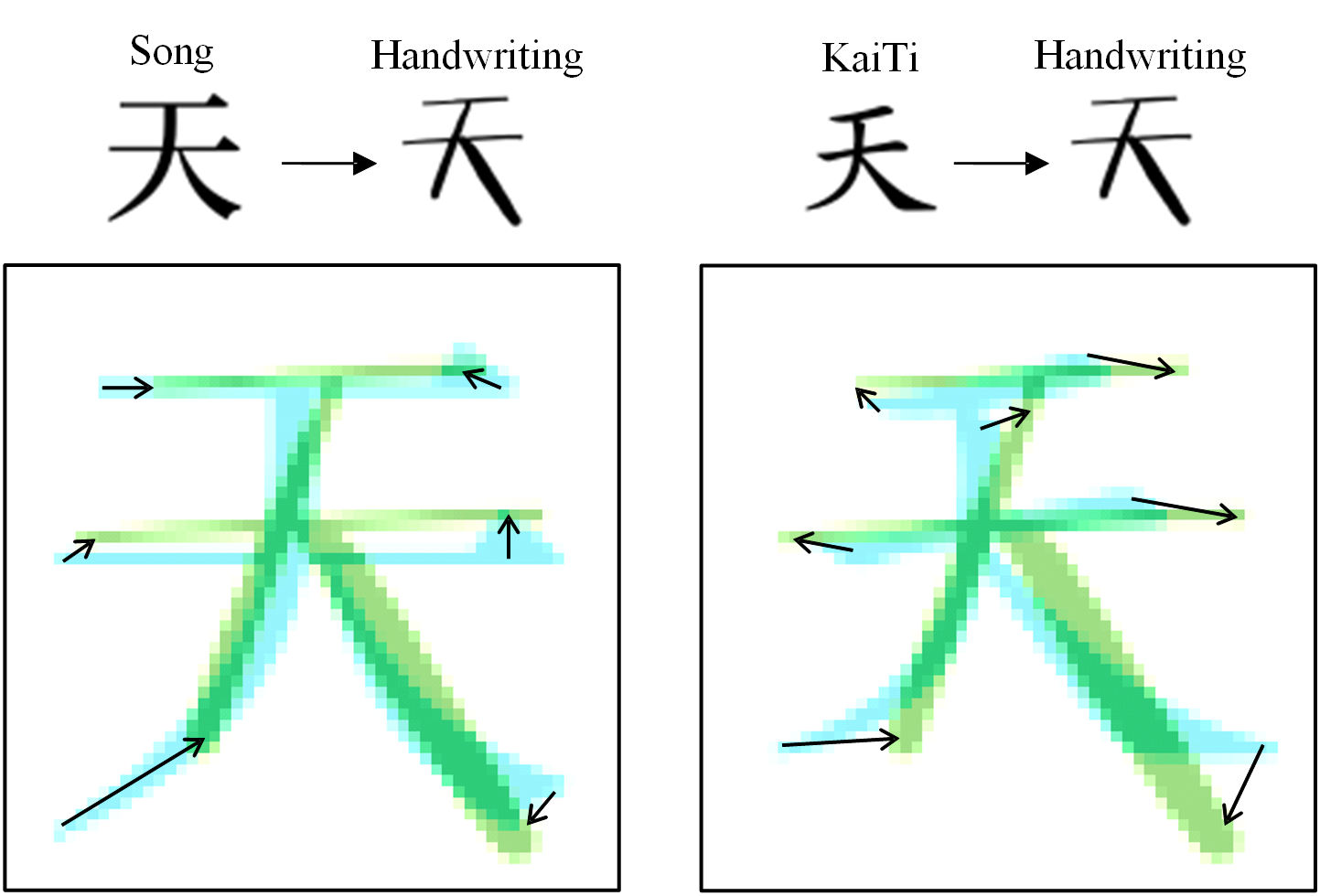}
\end{center}
   \caption{\textbf{The geometric deformation of two fonts for a character}. We employ the character ``Tian" to compare a handwritten style with the fonts of Kaiti and Song. There is a correspondence for each stroke between two fonts.}
\label{fig:3}
\end{figure}

As illustrated in Fig. \ref{fig:3}, there lies in a geometric deformation of two fonts for a character and exists a correspondence for each stroke. Compelled this observation, we propose a feature deformation skip connection (FDSC) module to apply geometric deformation convolution to the content image in the feature space and directly transfer the deformation low-level feature to the mixer. Specifically, the module predicts offsets based on the guidance code to instruct the deformable convolution layer performing a geometric transformation on the low-level feature. As demonstrated in Fig. \ref{fig:2}, the input of FDSC module is a concatenation of two feature maps: a feature map $K_c$ extracted from the content image and a style guidance map $K_s$. $K_s$ is extracted from the mixer after injecting the style code $Z_s$. The module estimates sampling parameters after applying convolution to the concatenation of $K_s$ and $K_c$:
\begin{equation}
\Theta = f_{\theta}(K_s, K_c).
\end{equation}
Here, $f_{\theta}$ refers to a convolution layer, and $\Theta$ = \{$\Delta p_{k}$, $\Delta m_{k}$ $|\  k$ = 1, $\cdots,|\mathcal{R}|$\} refers to the offsets and mask of the convolution kernel, where $\mathcal{R}$ = \{(-1, -1), (-1, 0), $\cdots$, (0, 1), (1, 1)\} indicates a regular grid of a 3$\times$3 kernel. Under the guidance of sampling parameter $\Theta$, a geometrically deformed feature map $K^{'}_{c}$ is obtained from $\Theta$ and $K_c$ based on deformable convolution ${f_{DC}(\cdot)}$:
\begin{equation}
K^{'}_{c} = f_{DC}(K_c, \Theta).
\end{equation}
Specifically, for each position $p$ on the output $K^{'}_{c}$, the deformable convolution $f_{DC}(\cdot)$ is applied as follow:
\begin{equation}
K^{'}_{c}(p) = \sum_{k=1}^{\mathcal{R}}w(p_k)\cdot x(p + p_k + \Delta p_k) \cdot \Delta m_k,
\end{equation}
where the $w(p_k)$ indicates the weight of the deformable convolution kernel at $k$-th location. The convolution is operated on the irregular positions ($p_k$ + $\Delta$ $p_k$) where $\Delta$ $p_k$ may be fractional.  Followed \cite{Deformablev1}, the operation is implemented by using bilinear interpolation. At last, the output of feature deformation skip connection module is fed to the mixer and  $K^{'}_{c}$ is then concatenated with $K_s$ like a common used skip connection  \cite{DBLP:conf/miccai/RonnebergerFB15}.

Deformable convolution introduces 2D offsets to the regular grid sampling locations in the standard convolution. It enables free form deformation of the sampling grid. There are lots of areas of the same color in character images, such as background color and character color. By using the deformable convolution, an area can be related to any other area with the same color. It is difficult to optimize the non-unique solution. To efficiently use our FDSC module, we impose a constrain on the offsets $\Delta p$. We introduce the constrain in detail in the next subsection. Section \ref{visualization} demonstrates the visualization of the offsets $\Delta p$.

Our FDSC module aims to deform the spatial structure of the content image in the feature space. It is crucial to select which level of features to be transformed. As we know, low-level features contain more spatial information than high-level features. In our model, we employ the feature maps after the first and second convolution layer as input to the FDSC module. Appendix B demonstrates the analysis of the performance of the model with different numbers of the FDSC module.

\subsection{Loss Function}

Our model aims to achieve automatic font generation via an unsupervised method. To this end, we adopt four losses: 1) adversarial loss is used to produce realistic images. 2)  content consistent loss is introduced to encourage the content of the generated image to be consistent with the content image; 3) image reconstruction loss is used to maintain the domain-invariant features; 4) deformation offset normalization is designed to prevent excessive offsets of the FDSC module. We introduce the formula of each loss and the full objective in this section.

\textbf{Adversarial loss: }the proposed network aims to generate plausible images by solving a minimax optimization problem. The generative network $G$ tries to fool discriminator $D$ by generating fake images. The adversarial loss penalty the wrong judgement when real/generated images are input to discriminator.
\begin{equation}
\begin{aligned}
\mathcal{L}_{adv} = \max_{D_s}\min_{G} \mathbb{E}_{I_s \in P_s, I_c \in P_c} [ \log D_s(I_s)
\\
+\log (1-D_s(G(I_s, I_c)))],
\end{aligned}
\end{equation}
where $D_s$($\cdot$) denotes the logit from the corresponding style of discriminator’s output.

\textbf{Content consistent loss: }adversarial loss is adopted to help the model to generate a realistic style while ignoring the correctness of the content. To prevent mode collapse and ensure that the features extracted from the same content can be content consistent after the content encoder $f_c$, we impose an content consistent loss here:
\begin{equation}
\label{Lcnt}
\mathcal{L}_{cnt} =  \mathbb{E}_{I_s \in P_s, I_c \in P_c}\left\|Z_c - f_c(G(I_s, I_c))\right\|_1 .
\end{equation}
${L}_{cnt}$ ensures that given a source content image $I_c$ and corresponding generated images, their feature maps are consistent after content encoder $f_c$.

\textbf{Image Reconstruction loss: } To ensure that the generator can reconstruct the source image $I_c$ when given with its origin style, we impose an reconstruction loss:
\begin{equation}
\label{Limg}
\mathcal{L}_{img} =  \mathbb{E}_{I_c \in P_c}\left\|I_c - G(I_c,I_c)\right\|_1 .
\end{equation}
The objective helps preserve domain-invariant characteristics (\eg, content) of its input image $I_c$.

\textbf{Deformation offset normalization: }
The deformable offsets enable free form deformation of the sampling grid.
As there are lots of areas of the same color between input images and generated images (such as background color and character color), it leads to a non-unique solution which is difficult to optimize. Meanwhile, the font generation focus on the stroke relationship between content character image and target character image, such as the thickness and tips of stroke. However, given images with the same content but different style, the position of the same stroke in these two images are close. To efficiently use this deformable convolutional network, we impose a constrain on the offsets $\Delta p$:
\begin{equation}
\mathcal{L}_{offset} =  \frac{1}{|\mathcal{R}|}\left\|\Delta p\right\|_1,
\label{offsetloss}
\end{equation}
where $\Delta p$ denotes offsets of the deformable convolution kernel, $|\mathcal{R}|$ denotes the number of the convolution kernel. 

\textbf{Overall Objective loss: }Combining all the above-mentioned loss, we have the overall loss function for training our proposed framework: 
\begin{equation}
\mathcal{L} = \mathcal{L}_{adv} +  \lambda_{img}\mathcal{L}_{img} + \lambda_{cnt}\mathcal{L}_{cnt} + \lambda_{offset}\mathcal{L}_{offset},
\end{equation}
where $\lambda_{adv}$, $\lambda_{img}$, $\lambda_{cnt}$, $\lambda_{offset}$ are hyperparameters to control the weight of each loss function.  In our model, the generative network aims to minimize the overall object loss, while the discriminator aims to maximize it.


\begin{table*}[t]
\begin{center}
\setlength{\tabcolsep}{11.5pt}{
\begin{tabular*}{17cm}{lcllllll}
\hline
Methods & one-to-many& training & L1 loss& RMSE& SSIM& LPIPS& FID \\
\hline
\multicolumn{8}{c}{\textbf{Seen} fonts } \\
\hline
EMD \cite{EMD}  & $\checkmark$ & paired &0.0538 &0.1955 &0.7676&0.1036 &89.65 \\
Zi2zi \cite{zi-2-zi} & $\checkmark$ & paired &\textbf{0.0521} &0.1802 &0.7789 & 0.1065 &142.23 \\
Cycle-GAN \cite{Zhu2017UnpairedIT}  & $\times$ & unpaired &0.0863 &0.2555 &0.6392 &0.1825 &175.24 \\
GANimorph \cite{ganimorph} & $\times$ & unpaired &0.0563 &\textbf{0.1759} &\textbf{0.7808} &0.1403 &72.89 \\
FUNIT \cite{DBLP:conf/iccv/0001HMKALK19} & $\checkmark$ & unpaired & 0.0807 &0.2510 &0.6669 &0.1216 &53.77 \\
Ours & $\checkmark$ & unpaired &0.0562 &0.1994 &0.7580 &\textbf{0.0814} & \textbf{46.15}\\
\hline
\multicolumn{8}{c}{\textbf{Unseen} fonts} \\
\hline
EMD \cite{EMD}  & $\checkmark$ & paired & 0.0430 &0.1755 &0.7849 &0.1255 &82.53 \\
FUNIT \cite{DBLP:conf/iccv/0001HMKALK19} & $\checkmark$ & unpaired &0.0588 &0.2089 &0.7417 &0.1125 &59.98 \\
Ours & $\checkmark$ & unpaired & \textbf{0.0414} &\textbf{0.1709} &\textbf{0.7982}& \textbf{0.0867}& \textbf{50.29}\\
\hline
\end{tabular*}}
\end{center}
\caption{\textbf{Quantitative evaluation on the whole dataset}. We evaluate the methods on seen and unseen font sets. The bold number indicates the best.}
\label{table1}
\end{table*}
    
\begin{figure*}
\centering
\subfigure[Easy cases (\ie, non-cursive writing).]{
\begin{minipage}[b]{1\textwidth}
\includegraphics[width=0.99\textwidth]{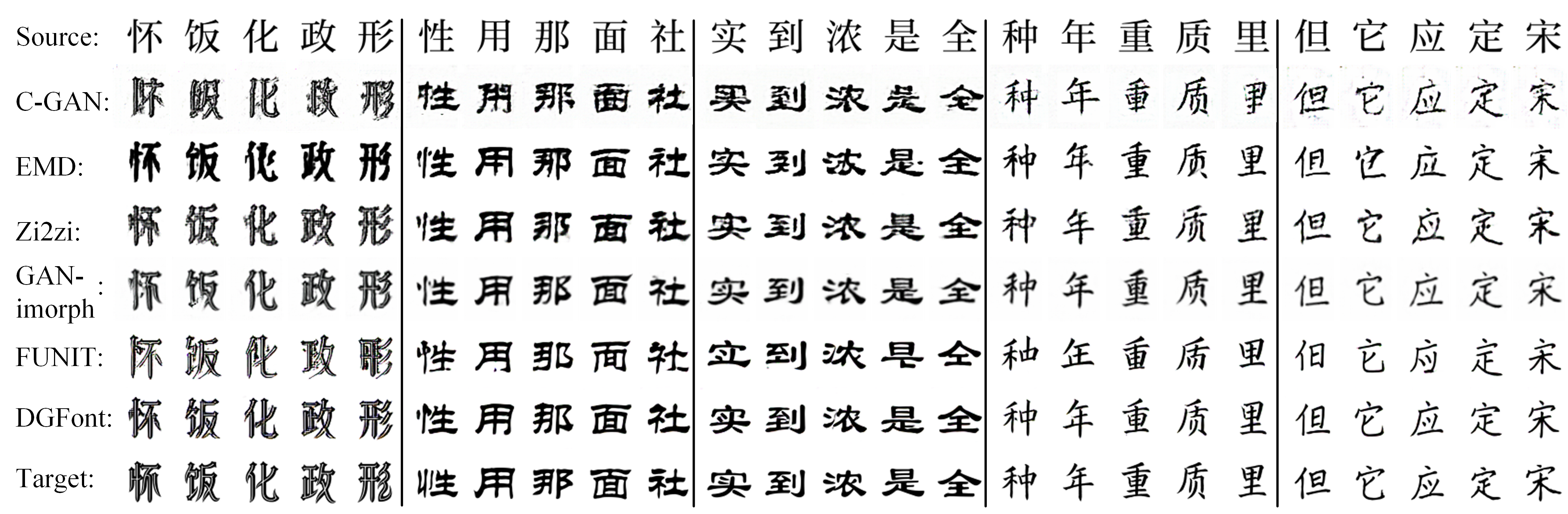}\label{fig:4a}
\end{minipage}
}
\subfigure[Challenging cases (\ie, cursive writing).]{
\begin{minipage}[b]{1\textwidth}
\includegraphics[width=1\textwidth]{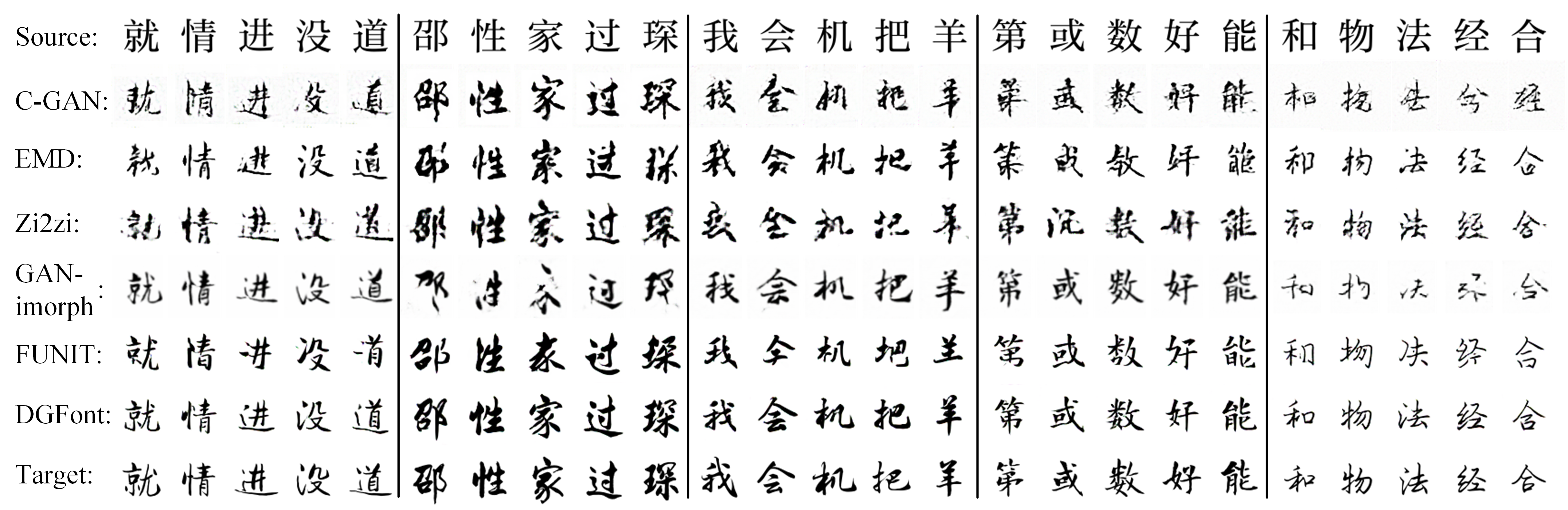}\label{fig:4b}
\end{minipage}
}
\caption{Comparisons to the stat-of-art methods for font generation.} \label{fig:results}
\label{reuslts}
\end{figure*}

\section{Experiments}
In this section, we evaluate our proposed model for the Chinese font generation task. We first introduce our dataset. After that, the results of our experiments are shown to verify the advantages of our model. More implementation details are shown in Appendix A.

\subsection{Dataset}
To evaluate our model for Chinese font generation, we collect a dataset that contains 410 fonts (styles) including handwritten fonts and printed fonts, each of which has 990 commonly used Chinese characters (content). All images are 80$\times$80 pixels. The dataset is randomly partitioned into a training set and testing set. The training set contains 400 fonts, and each font contains 800 characters. The testing set consists of two parts. One part is the remaining 190 characters of the 400 fonts. Another part is the remaining 10 fonts which are used to test the generalization ability to unseen fonts.
\subsection{Comparison with State-of-art Methods}
In this subsection, we compare our model with the following methods for Chinese font generation: 1) Cycle-GAN \cite{Zhu2017UnpairedIT}: Cycle-GAN consists of two generative networks which can translate images from one domain to another using a cycle consistency loss. Cycle-GAN is also an unsupervised learning method; 2) EMD \cite{EMD}: EMD employs an encoder-decoder architecture, and separates style/content representations. EMD is optimized by L1 distance loss between ground-truth and generated images; 3) Zi2zi \cite{zi-2-zi}: Zi2zi is a modified version of pix2pix \cite{Isola2017ImagetoImageTW} model, it achieves font generation and uses Gaussian Noise as category embedding to achieve multi-style transfer. Zi2zi still requires paired data; 5) GANimorph \cite{ganimorph}: GANimorph adopts the cyclic image translation framework like Cycle-GAN and introduce a discriminator with dilated convolutions to get a more context-aware generator; 6) FUNIT \cite{DBLP:conf/iccv/0001HMKALK19}: FUNIT is an unsupervised image-to-image translation model which separates content and style of natural animal images and combine them with adaptive instance normalization (AdaIN) layer.

\begin{figure}[t]
\begin{center}
   \includegraphics[width=1\linewidth]{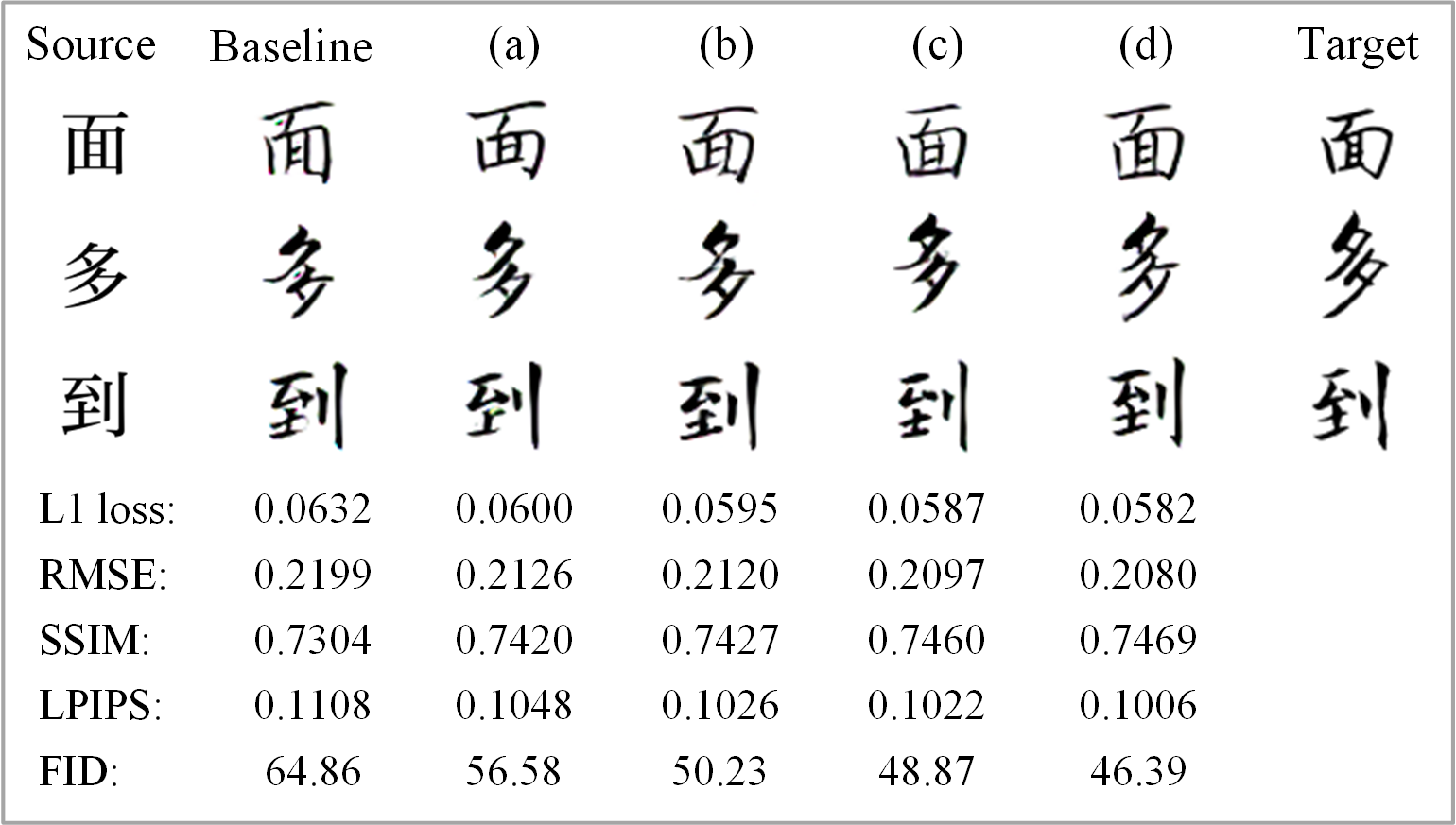}
\end{center}
   \caption{\textbf{Effect of different components in our method}. We add different parts into our baseline successively. (a) Replace the first three convolution layers of content encoder with deformable convolution layers; (b) add the FDSC-1 module (without normalization); (c) impose normalization on FDSC-1 module; (d) add the FDSC-2 module (full model).}
\label{fig:ablation}
\end{figure}
For a fair comparison, we employ the font of Song as the source font which is commonly used in font generation task \cite{EMD,RD-GAN}. Our model, EMD, TUNIT are trained with 400 fonts. For Cycle-GAN and GANimorph, they can only train one paired translation at once, hence we train 399 models of Cycle-GAN individually for each target style. In our experiments, we find that the model of Zi2zi trained with 400 fonts performs worse than trained with two fonts. As a result, we train 399 models for Zi2zi for each target style.

\textbf{Quantitative comparison}.
The quantitative results are shown in Table \ref{table1}. In the experiments, DG-Font is comparable to compared methods in pixel-level evaluation metrics, \eg, L1 loss, RMSE, SSIM. It is noted that these metrics focus on pixel-wise between generated image and ground-truth and ignore the feature similarity which is closer to human perceptions. In perceptual-level metrics FID \cite{DBLP:conf/nips/HeuselRUNH17} and LPIPS \cite{DBLP:conf/cvpr/ZhangIESW18}, we can observe that DG-Font outperforms the compared methods and reaches the state-of-the-art performance for both seen fonts and unseen fonts. 



\textbf{Qualitative comparison}.
In order to verify the capability of deforming and transforming source character patterns (\eg, stroke, skeleton), two kinds of visual comparisons are displayed in Fig. \ref{reuslts}. First, we compare DG-Font to other methods with relative simple fonts that are close to printed fonts with no cursive writing. As demonstrated in Fig. \ref{fig:4a}, Cycle-GAN can only generate parts of characters or sometimes unreasonable structures. Characters generated by Zi2zi EMD, and GANimorph can maintain a complete structure, but they are usually vague. FUNIT can generate characters with a clear background but the generated characters lose their structure to some degree. DG-Font is able to generate character close to the target well. In contrast to fonts in Fig. \ref{fig:4a}, fonts in Fig. \ref{fig:4b} are more challenging for the rich details and joined-up writing. We can observe that Cycle-GAN, EMD, Zi2zi and GANimorph can hardly generate characters under challenging cases. While FUNIT maintains the ability to generate characters with incomplete structure, but the skeleton of generated character is not well transformed. Our proposed DG-Font can not only generate characters with complete structure but also learn joined-up writing.

\subsection{Ablation Study}\label{ablation}
\begin{table}[]
\begin{center}
\begin{tabular}{llllll}
\hline
Method                   & L1 loss & RMSE & SSIM & LPIPS & FID \\ \hline
SC &    0.0641     &   0.2212   &   0.7252   &    0.1114   &  46.88   \\
FDSC                     &\textbf{0.0582}     &\textbf{0.2080}    & \textbf{0.7469}     &    \textbf{0.1006}   &  \textbf{46.39}   \\
 \hline
\end{tabular}
\end{center}
\caption{\textbf{Comparison with skip-connection (SC) proposed by U-Net} \cite{DBLP:conf/miccai/RonnebergerFB15} . We replace two FDSC modules with skip-connections and then compare the new model with the full model of DG-Font.
        }
\label{unet}
\end{table}


In this part, we add different parts into the model successively and analyze the influence of each part, including deformable convolution, feature deformation skip connection and deformable offset normalization. We conduct the ablation study on the data set of 187 handwritten fonts. Our baseline is the models that replace deformable convolution with normal convolution and remove FDSC modules. 
Qualitative and quantitative comparisons are shown in Fig. \ref{fig:ablation}.

1) \textbf{Effectiveness of deformable convolution in the content encoder.} Fig. \ref{fig:ablation}(a) shows the results by replacing the first three convolution layers of the content encoder with deformable convolution layers. We can see that the quantitative results improve obviously in terms of L1 loss, RMSE, and SSIM. This indicates that deformable convolution layers in the content encoder effectively help improve the performance of our model.

2) \textbf{The influence of the FDSC module.} In this part, we add an FDSC module (without offset normalization in Eq. \ref{offsetloss}) that connects the features after the first layer and penultimate layer. Results are shown in Fig. \ref{fig:ablation}(b). Comparing with Fig. \ref{fig:ablation}(a), we observe that the generated characters preserve more structure information and are able to reconstruct the complete structure of characters. 

3) \textbf{Effectiveness of deformable offset constrain.} We investigate the impact of deformable offset normalization by comparing FDSC module without and with offset normalization. As shown in Fig. \ref{fig:ablation}(b) and (c), adding offset normalization helps the model generate images whose style become more similar to the target.

4) \textbf{Effectiveness of two FDSC modules.} Fig. \ref{fig:ablation} (d) shows the results of our full model with two FDSC modules. It is noted that the generated images get more details, less noise, and achieves better quantitative results.

In addition, we compare our proposed FDSC module with common used skip-connection \cite{DBLP:conf/miccai/RonnebergerFB15, EMD, SCFont, CalliGAN} proposed by U-Net \cite{DBLP:conf/miccai/RonnebergerFB15}. Skip-connection is often adopted to transfer feature maps with different resolution directly from encoder to decoder, which is effective in semantic segmentation \cite{DBLP:conf/cvpr/JegouDVRB17,DBLP:conf/cvpr/LongSD15} and photo-to-art \cite{DBLP:journals/corr/abs-2003-07694} tasks whose content of inputs and outputs share the same structure. However, the font generation requires a geometric deformation between content inputs and the corresponding generated images in structure. To compare FDSC module with skip-connection, We replace two FDSC modules with skip-connection in our proposed DG-Font network. The comparison results are shown in Table \ref{unet}. We can observe that models with FDSC modules outperform models with skip-connection, which prove the advantage of FDSC.

\begin{figure}[t]
\begin{center}
   \includegraphics[width=\linewidth]{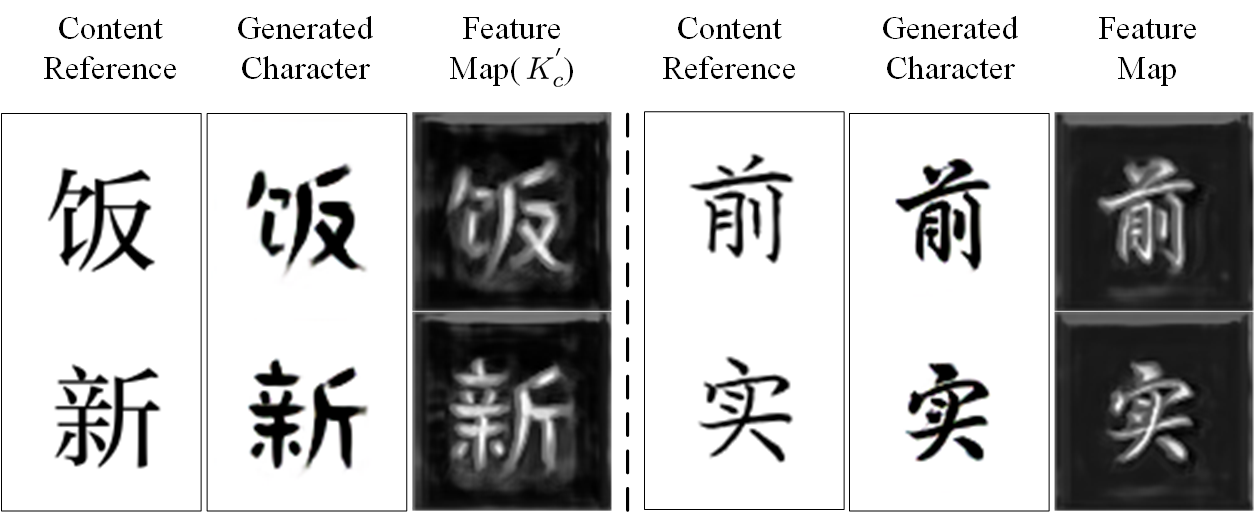}
\end{center}
   \caption{\textbf{Feature visualization}. We visualize the features $K^{'}_{c}$ generated from the FDSC-1 module. For each case, from left to right: content reference characters, the corresponding generated characters, the visualization of feature maps. For feature map images, the whiter the area, the larger the activation value.}
\label{fig:5}
\end{figure}

\begin{figure}[t]
\begin{center}
   \includegraphics[width=0.9\linewidth]{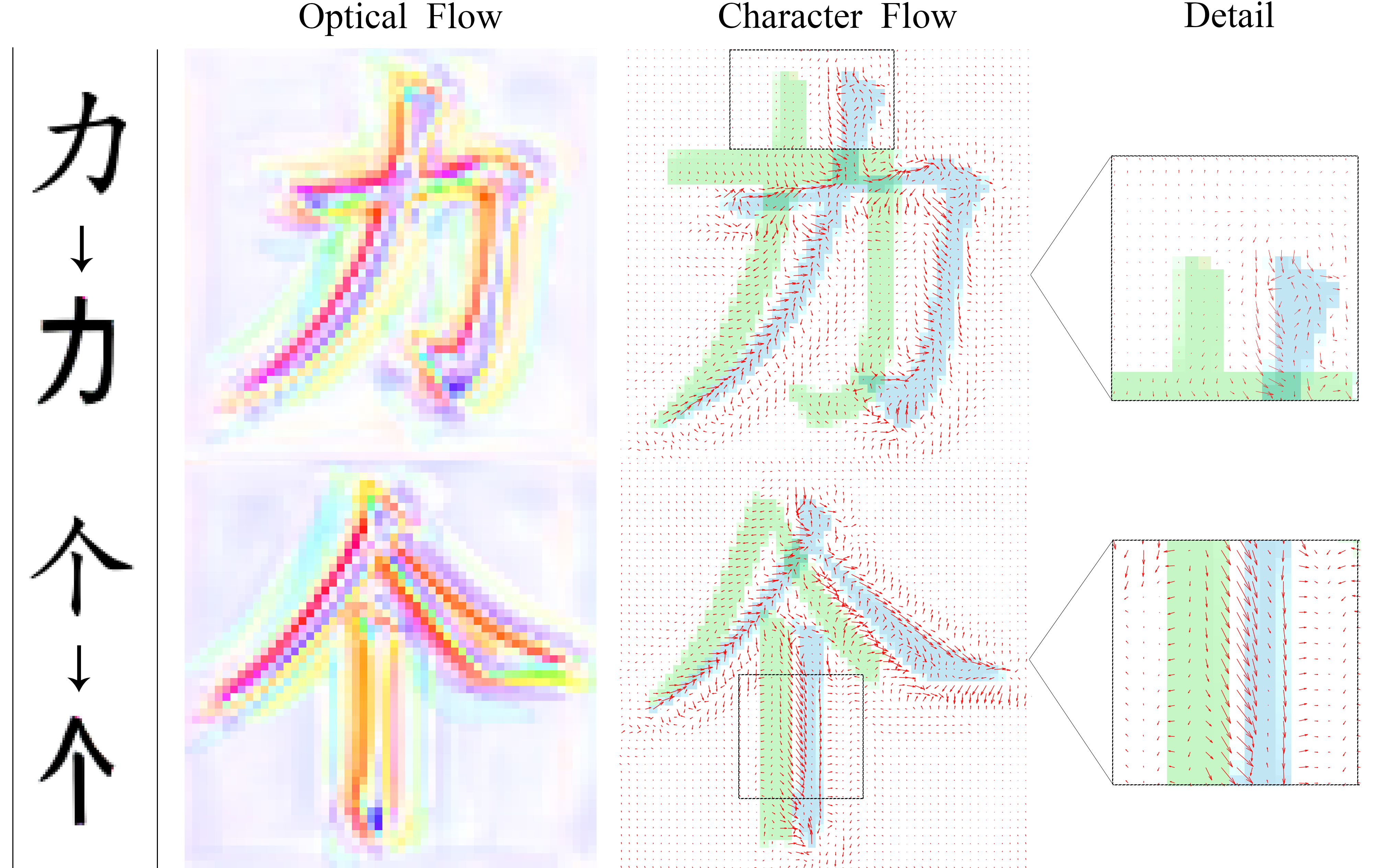}
\end{center}
   \caption{\textbf{The visualization of learned offsets}. First column: source image and generated image. Second column: the optical flow displays the estimated offsets $\Delta p$. Third column: character flow visualized the offsets $\Delta p$. Forth column: zoomed-in details. Source and generated images are in blue and green respectively.}
\label{fig:6}
\end{figure}

\subsection{Visualization}\label{visualization}
In order to show the effectiveness of FDSC, we visualize the feature maps generated by the FDSC-1 module. As shown in Fig. \ref{fig:5}, the feature maps $K^{'}_{c}$ preserve the pattern of characters well, which helps generate a character with complete structure. On the other hand, we can observe that the FDSC module effectively transform features extracted from the content encoder.

In addition, we visualize the learned offsets from the FDSC-1 module using optical flow and character flow respectively. To visualize the offsets clearly, the kernel of deformable convolution in the FDSC module is set to 1$\times$1.
As demonstrated in Fig. \ref{fig:6}, we observe that the learned offsets mainly affect the character region. The offsets value of the background tends to zero, which proves the usefulness of the proposed offset loss Eq. \ref{offsetloss}. In character flow, we can see that most of the offset vectors point from the stroke in target characters to the corresponding source stroke. The results show that in the convolution process, the sampling locations of target characters tend to shift to corresponding locations in source character by the learned offsets.


\section{Conclusion}
In this paper, we propose an effective unsupervised font generation model which is capable to generate realistic characters without paired images and can extend to unseen font well. To ensure the integration of generated characters, we propose a Feature Deformation Skip Connection (FDSC) module to transfer the deformable low-level spatial information to the mixer. Besides, we employ deformable convolution layers in content encoder to learn style-invariant feature representations. Extensive experiments on Chinese font generation verify the effectiveness of our proposed model.

\noindent \textbf{Acknowledgements}
This work was partly supported by the National Key Research and Development Program of China under No. 2020AAA0107903, the China Postdoctoral Science Foundation under No. 2020M681237, and the Science and Technology Commission of Shanghai Municipality under No.19511120800,  No.19ZR1415900, No.18DZ2270800.

{\small
\bibliographystyle{ieee_fullname}
\bibliography{egbib}
}

\newpage

\section*{Appendix}

\subsection*{A. Implementation Detail}


\subsubsection*{A.1 Training Strategy}
We initial the weights of convolutional layers with He initialization, in which all biases are set to zero and the weights of linear layers are sampled from $N(0, 0.01)$. We use Adam optimizer with $\beta_1$ = 0.9 and $\beta_2$ = 0.99 for style encoder, and RMSprop optimizer with $\alpha$ = 0.99 for the content encoder and mixer. We train the whole framework 200K iterations and the learning rate is set to 0.0001 with a weight decay 0.0001. We train the model with a hinge version adversarial loss with R1 regularization using $\gamma=10$. In all experiments, we use the following hyper-parameters: $\lambda_{img}$ = 0.1, $\lambda_{cnt}$ = 0.1, $\lambda_{offset}$ = 0.5. All the images are resized to 80$\times$80 before training and testing and the batch size is set to 32. In testing process, we use ten reference images to compute an average style code for the generation process. The source code will be released after the paper is accepted.

\subsubsection*{A.2 Network Architecture}
The proposed model is an encoder-decoder network. The style encoder whose architecture is based on VGG-11 aims to extract style information. The architecture of the content encoder and mixer are symmetrical, which help preserve domain-invariant information of content. The detailed architectures of style encoder, content encoder, and mix are shown in Table \ref{architecture1}. The detailed information of discriminator is listed in Table \ref{architecture2}.

\subsection*{B. Ablation Study}
We evaluate our model by setting different hyper-parameter values in overall objective loss (Eq. 8). All the metrics are computed based on the 400 seen fonts mentioned in Sec 4.1.

\textbf{ Content reconstruction loss}: We analyze the impact of the content reconstruction loss in Table \ref{t1}. We observe that a large $\lambda_{cnt}$ value leads to degradation, while the model begins even worse without the content reconstruction loss. It show that $\lambda_{cnt}$ = 0.5 provides a good trade-off.

\textbf{Image reconstruction loss:}  As shown in Table \ref{t1}, we find that our method performs well when $\lambda_{img}$ = 0.1. The image reconstruction loss helps preserve domain-invariant characteristics of the content input, but a large $\lambda_{img}$ makes the model pay more attention to reconstruct input images and perform poorly on generating new characters.

\textbf{Offset normalization: }Table \ref{t1} presents results of loss term $\lambda_{offset}$ in deformable offset normalization (Eq. 7).  As demonstrated in the table, $\lambda_{offset}$ = 0.5 significantly improve the generation quality.

\textbf{Influence of the number of FDSC module: }to evaluate the influence of the number of FDSC module on model performance, we conduct experiments that adding FDSC module transfer different levels of information from low-level to high level sequentially. From Table \ref{t4}, we can observe that adding an FDSC module to transfer the deformed low-level feature significantly improve the performance of the model. However, when the second FDSC module is added to transfer the higher-level information, the improvement is not obvious. This may be because feature maps input to the second FDSC contains less spatial information. The third FDSC even reduces performance. The detailed results are shown in Table \ref{t4}.

\begin{table}[t]
\begin{center}
\begin{tabular}{cccccc}
\hline
                   & L1 loss & RMSE & SSIM & LPIPS & FID \\ \hline
N = 1   &     0.0563    &   0.2005   &  0.7549    &  0.850     &   47.99  \\
N = 2 &    \textbf{0.0562}     &    \textbf{0.1994}  &  \textbf{0.7580}   &  \textbf{0.0814}     &    \textbf{46.15} \\
N = 3   &    0.0580     &   0.2039   &  0.7514   &    0.0832  &   47.12  \\ \hline
\end{tabular}
\end{center}
\caption{\textbf{The impact of the number of FDSC module}. $N$ denotes the number of FDSC module. }
\label{t4}
\end{table}



\subsection*{D. More Results}
\begin{figure*}[ht]
\begin{center}
\includegraphics[width=1\textwidth]{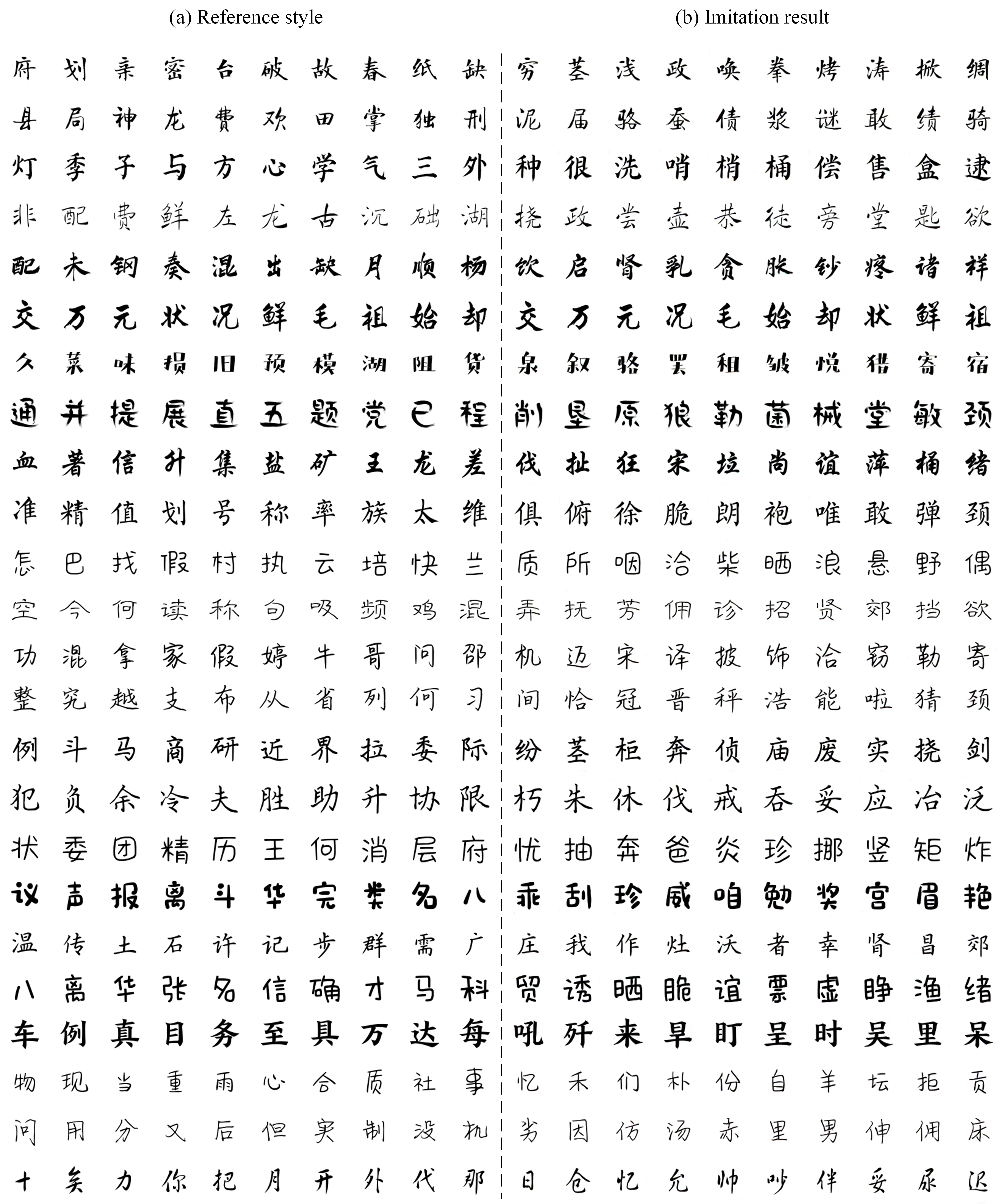}
\end{center}
   \caption{\textbf{Our results of different fonts on each raw}. For each case, the left ten columns show the style fonts, and the right ten columns show the corresponding generated imitation results.}
\label{more_results}
\end{figure*}

We select twenty-four fonts including calligraphy and handwriting to prove the superiority of our method. These fonts present different styles in geometric transformation, stroke thickness, tips, and joined-up writing patterns. The results show that the proposed method outputs high-quality characters. The results are shown in Fig. \ref{more_results}.

\newpage

\begin{table*}[ht]
\begin{center}
\setlength{\tabcolsep}{9mm}{
\scalebox{0.9}{
\begin{tabular}{cccccc}
\hline
                         & L1 loss & RMSE   & SSIM   & LPIPS  & FID   \\ \hline
$\lambda_{cnt} = 0$      & 0.0592  & 0.2065 & 0.7441 & 0.0874 & 52.91 \\
$\lambda_{cnt} = 0.1$    & \textbf{0.0562}  & \textbf{0.1994}   &  \textbf{0.7580}  & \textbf{0.0814} &\textbf{46.15}  \\
$\lambda_{cnt} = 1$      &   0.0591      &   0.2068     &   0.7456     &  0.0849      &  46.60     \\ \hline
$\lambda_{img} = 0$   &     0.0593    &   0.2065   &  0.7439    &    0.0860   &  59.29   \\
$\lambda_{img} = 0.1$ &    \textbf{0.0562}     &  \textbf{0.1994}    &  \textbf{0.7580}    &  \textbf{0.0814}      & 46.15   \\
$\lambda_{img} = 1$   &    0.0627     &   0.2155   &  0.7340    &   0.0908    &   \textbf{44.86}  \\ \hline
$\lambda_{offset} = 0$   &    0.0568     &    0.2007  &   0.7532   &    0.0870   &   52.60  \\
$\lambda_{offset} = 0.5$ &    \textbf{0.0562}     &    \textbf{0.1994}  &  \textbf{0.7580}   &  \textbf{0.0814}     &    \textbf{46.15} \\
$\lambda_{offset} = 1$   &    0.0569     &    0.2006  &  0.7538    &    0.0838   &   50.23  \\ \hline
\end{tabular}}}
\end{center}
\caption{\textbf{Impact of the hyper-parameters}. }
\label{t1}
\end{table*}

\begin{table*}[ht]
\begin{center}
\scalebox{0.9}{
\begin{tabular}{cccccccc}
\hline
\hline
                                 & Operation                              & Kernel size & Resample & Padding & Feature maps & Normalization & Nonlinearity \\ \hline
\multirow{8}{*}{Style encoder}   & Convolution                            & 3           & MaxPool  & 1       & 64           & BN            & ReLU         \\
                                 & Convolution                            & 3           & MaxPool  & 1       & 128          & BN            & ReLU         \\
                                 & Convolution                            & 3           & - & 1       & 256          & BN            & ReLU         \\
                                 & Convolution                            & 3           & MaxPool  & 1       & 256          & BN            & ReLU         \\
                                 & Convolution                            & 3           & - & 1       & 512          & BN            & ReLU         \\
                                 & Convolution                            & 3           & MaxPool  & 1       & 512          & BN            & ReLU         \\
                                 & Convolution                            & 3           & - & 1       & 512          & BN            & ReLU         \\
                                 & Convolution                            & 3           & MaxPool  & 1       & 512          & BN            & ReLU         \\ 
                                 & Average pooling                                     & -           & -        & -       & 128          & -             &  -   \\
                                 & FC                                     & -           & -        & -       & 128          & -             &  -   \\
                                 \hline
                                 
\multirow{4}{*}{Content encoder} & Deformable conv                        & 7           & - & 3       & 64           & IN            & ReLU         \\
                                 & Deformable conv                        & 4           & stride-2 & 1       & 128          & IN            & ReLU         \\
                                 & Deformable conv                        & 4           & stride-2 & 1       & 256          & IN            & ReLU         \\
                                 & Residual block $\times$ 2 & 3           & - & 1       & 256          & IN            & ReLU         \\ \hline
                                 
\multirow{4}{*}{Mixer}           & Residual block $\times$ 2 & 3           & - & 1       & 256          & AdaIN         & ReLU         \\
                                 & Convolution                            & 5           & Upsample & 2       & 128          & AdaIN         & ReLU         \\
                                 & Convolution                            & 5           & Upsample & 2       & 64           & AdaIN         & ReLU         \\
                                 & Convolution                            & 7           & - & 3       & 3            & -             & tanh       \\ \hline
\end{tabular}}
\end{center}
\caption{\textbf{Generative network architecture}. BN, IN, AdaIN denote the batch normalization, Instance normalization, and Adaptive instance normalization, respectively. FC means the fully connected layer}
\label{architecture1}
\end{table*}

\begin{table*}[ht]
\begin{center}
\scalebox{0.9}{
\begin{tabular}{cccccccc}
\hline
                                & Operation      & Kernel size & Resample & Padding & Feature maps & Normalization & Nonlinearity \\ \hline
\multirow{11}{*}{Discriminator} & Convolution    & 3           & - & 1       & 64           & -             & -            \\
                                & Residual block & 3           & - & 1       & 64           & FRN           & -            \\
                                & Residual block & 3           & AvgPool  & 1       & 128          & FRN           & -            \\
                                & Residual block & 3           & - & 1       & 128          & FRN           & -            \\
                                & Residual block & 3           & AvgPool  & 1       & 256          & FRN           & -            \\
                                & Residual block & 3           & - & 1       & 256          & FRN           & -            \\
                                & Residual block & 3           & AvgPool  & 1       & 512          & FRN           & -            \\
                                & Residual block & 3           & - & 1       & 512          & FRN           & -            \\
                                & Residual block & 3           & AvgPool  & 1       & 1024         & FRN           & LeakyReLU    \\
                                & Convolution    & 4           & - & 1       & 1024         & -             & LeakyRuLU    \\
                                & Convolution    & 1           & AvgPool  & 1       & 400          & -             & -            \\ \hline
\end{tabular}}
\end{center}
\caption{\textbf{Discriminator architecture}. AvgPool denotes the averge pooling. The slope of LeakyRuLU is set to 0.2.}
\label{architecture2}
\end{table*}

\end{document}